\documentclass{article}
\pdfoutput=1

    \usepackage{neurips_2020}
\usepackage[utf8]{inputenc} 
\usepackage[T1]{fontenc}    
\usepackage{hyperref}       
\usepackage{url}            
\usepackage{booktabs}       
\usepackage{amsfonts}       
\usepackage{nicefrac}       
\usepackage{microtype}      
\usepackage{graphicx}       

\graphicspath{ {./images/} }

\title{Advancing Humor-Focused Sentiment Analysis through Improved Contextualized Embeddings and Model Architecture}

\author{
}

\begin{document}

\nolinenumbers

\maketitle 

\section{Introduction and Motivation}

Humor is a natural and fundamental component of human interactions. Broadly, linguists recognize humor as the substitution of the literal meaning of words in a sentence by a figurative meaning, usually contradictory with the original meaning, to express feelings and ideas. This is evident in sentences such as “I love when I miss the bus in the morning”. To us, humans, it is natural to classify this sentence as humor – or more specifically sarcasm – because it is obvious that one does not truly appreciate missing the bus. When correctly applied, humor allows us to express thoughts and feelings conveniently and effectively, increasing interpersonal affection, likeability, and trust. However, understanding the use of humor is a computationally challenging task from the perspective of humor-aware language processing models. Computational models do not intrinsically possess the prior world knowledge humans employ to perform the classification task of a sentence as a humorous attempt, therefore researchers must design elaborate solutions to mathematically express nonliteral meaning effectively enough for a model to recognize it as a feature. 

As language processing models become ubiquitous through virtual-assistants and IOT devices, the need to develop humor-aware models rises exponentially. The development of social robots, a long-lasting desire of researchers in Artificial Intelligence and Robotics, is an example of a research field currently limited by insufficient performance on humor-focused sentiment analysis. Social robots are envisioned to be a reliable tool in the therapeutical intervention for children with autism spectrum disorder and be friendly companions and assistants to elders and younglings (\cite{Ritschel}). This creates the need for social robots to interact with humans in a very fluid manner, creating meaningful and genuine connections.  To achieve this goal, social robots must effectively detect and adapt to a myriad of human inputs, including their linguistic behavior and use of humor in interactions through nonliteral meanings. Humor becomes a key step to this field as it is a fundamental aspect to make interactions more natural, enjoyable, and to increase credibility and acceptance (\cite{Ritschel}) between humans and artificial entities such as robots and virtual assistants.

The current performance of humor-focused sentiment-analysis tasks is capped by the underlying way in which humans demonstrate to each other the usage of nonliteral meanings. First, humor is extremely context dependent. The sentence “it is a beautiful day outside” can carry literal or nonliteral meaning depending on weather experienced that day, and even so, the definition of a “beautiful day” is heavily personal, leading to contradictions among some humans. A model thus provided exclusively with a text input for “it is a beautiful day outside” cannot be reasonably expected to detect use of humor and other nonliteral meanings reliably. Additionally, humans convey nonliteral meanings through the use of both verbal and nonverbal behaviors during face-to-face communication, often varying vocal patterns and facial expressions (\cite{wang}). Consider the sentence “I am feeling great today” delivered by a child to a therapy robot. A human therapist would identify verbal and physical signs of distress to assess the veracity of this statement, potentially indicating nonliteral meaning being consciously or subconsciously employed by the child. Without feeding this information to a computational model, it is again expected to underperform human expectations.

For these reasons, early approaches in the field of humor-focused and broad nonliteral sentiment-analysis were unable to capture the subtle forms of context incongruity and nonverbal clues which lie at the heart of human communication (\cite {Joshi}). To further improve the state-of-the-art capacity to perform this particular sentiment-analysis task we must explore models that incorporate these contextualized and nonverbal elements in their design. Ideally, we seek architectures accepting these elements as additional embedded inputs to model, alongside the original sentence-embedded input. This survey thus analyses the current state of research in techniques for improved contextualized embedding incorporating nonverbal information, as well as newly proposed architectures to improve context retention on top of popular embeddings methods such as GloVe (\cite{pennington}) and Word2Vec (\cite{mikolov}). Therefore, the assembled collection of papers introduces new ideas on both fronts of this effort: improving nonverbal contextualized embeddings, and designing more humor-aware models for current embeddings. Currently there are no available comprehensive surveys specifically addressing and comparing these new models in the realm of understanding humor in informal conversation, and this paper seeks to fill this gap.

\section{Methodology}

Researchers confronting the humor-focused sentiment-analysis issue widely agree that current embedding procedures in pre-processing steps are not sufficient for modeling highly dynamic multimodal language. All studies analyzed in this paper have proposed novel techniques relating to improved contextualized embeddings for inputs, but they differ greatly in their motivation and design philosophy. Neural approaches are more frequent, however non-neural approaches are not rare. The trade-off between these approaches is a known dilemma in the field of Machine Learning. Given a certain complex-enough computational learning task to be performed, most frequently a neural architecture has an advantage on non-neural models through its ability to tune its parameters to capture more complex relationships between features. Evidently, state-of-the-art language models such as BERT and GPT-3 are neural architectures, which leverage on over $110$ million and $175$ billion trainable parameters, respectively. This additional performance however comes at the clear cost of increased computational complexity, and on its turn, increased computational complexity leads to higher development cost and training times, as it requires exponentially larger amounts of resources. The developers of GPT-3 at OpenAI calculated that training this particular model on a NVIDIA Tesla V100 GPU cloud environment (advertised as "the most advanced data center GPU ever built to accelerate AI") would require $355$ years and cost approximately $\$4.6$ million dollars. Presented with this fact, it becomes clear why some researchers choose to - or only have resources to - explore non-neural approaches. Although demonstrating more limited potential against neural networks due to possessing dozens of parameters instead of million of parameters, these Machine Learning models require low enough amounts of resources that even consumer laptops are capable of developing them locally. This makes experimentation on non-neural models extremely simple and accessible, so it is very important that researchers also seek to design efficient non-neural models, as they are more democratic and cost-effective solutions to ultimately be deployed on consumer products.    

To offer a comparison between the performance and success of neural against non-neural designs, we shall start this analysis with non-neural approaches, and later move into the realm of deep learning architectures. The paper “Sarcastic or Not: Word Embeddings to Predict the Literal or Sarcastic Meaning of Words” (\cite {Ghosh}) is a good initial point for this analysis, as it presents the most intuitive and simplistic approach. The authors propose an SVM-based approach considering a reframing of the nonliteral meaning detection task as a disambiguation task between literal meaning and sarcastic meaning. The Literal / Sarcastic Sense Disambiguation (LSSD) task considers every word in the sentence to be the potential target word for disambiguation, call it $t$, then produces two context-vectors to represent potential senses of $t$: sarcastic sense or literal sense. These vectors are referred to as $v_s$ and $v_l$, and they correspond to embeddings learned during the supervised training process. During validation, a sentence is provided to the model, which for every word in the sentence fetches its $v_s$ and $v_l$ representations developed in training, and produces a $v_u$ by embedding the target word on-the-fly using current state-of-the-art embedding procedures such as Word2Vec or GloVe. These three vectors are then combined to create a unified word-level embedding, and the model applies cosine similarity to the vector pairs $<v_u,v_s>$ and $<v_u,v_l>$ to score the likelihood of $t$ demonstrating sarcastic meaning. An interesting consequence of this design is the ability to extract word-specific performance, which demonstrates that the words $\{$“love”, “joy”, “brilliant”, “cute”$\}$ are the easiest to classify as sarcastic under the LSSD task, while $\{$“wonder”, “nice”, “interested”$\}$ create the highest difficulty for the model. This can be explained by considering that words such as $\{$“joy”, “cute”, “brilliant”$\}$ demonstrate considerably different contexts for literal and sarcastic usages, and word embeddings are able to capture and exploit these nuance distinctions (\cite {Ghosh}). On macro levels, it is remarkable that this rather simple approach achieves a 7-8\% increase in F1 performance when compared to presented baseline performance, which does not incorporating the additional proposed embedding procedure. This result is also independent of the embedding technique applied to obtain $v_u$.

A similar approach was attempted by Joshi et al. in the 2016 paper “Are Word Embedding-based Features Useful for Sarcasm Detection?”. Their goal was to experiment with word-embedding techniques to capture nonliteral meaning through context incongruity in the absence of sentiment words. The authors were motivated by a hypothesis that previous work relied excessively on patterns of sentiment-bearing words such as “love” and “hate” which may not capture nuanced forms of humor. Essentially, they introduce a concern for current models susceptibility to over-fit on the expectation of presence of sentiment words in nonliteral communication. Dr. Joshi and his colleagues thus proposed a reframing of the nonliteral meaning classification task from the linguistic perspective, suggesting word-level similarity or discordance to be an indicative of context incongruity, and a clue to humorous attempts. A famous quote from Australian writer Irina Dunn is presented as an example: “a woman needs a man like a fish needs a bicycle”. None of the words in this sentence have strong sentiment-bearing meaning, therefore initial work in this field would be biased towards a literal classification. Under the technique presented in the paper, however, we apply cosine-similarity to the vector-embeddings of every word pair, and the strong similarity for $<$“man”, “woman”$>$ $(s=0.766)$ and low similarity for $<$“fish”, “bicycle”$>$ $(s=0.131)$ indicates semantic discordance and this context incongruity is a clue to humorous intent.

In general, the technique consists of capturing the most similar and most dissimilar word pairs in the sentence, and use their scores as features for the sarcasm detection task. Each sentence is therefore augment with four features: maximum score of most similar word pair, maximum score of least similar word pair, minimum score of most similar word pair, minimum score of least similar word pair. It is also advised to divide the similarity scores by the square root of distance between the two words, so that closer pairs in the sentence are weighted higher, as that more likely indicates a contradictory meaning. Since such semantic similarity is but one of the components of context incongruity and existing feature sets have achieved respectable performance relying on sentiment-based features, it is imperative that the two be combined for the detection task (\cite {Joshi}). To validate the proposal, the authors conduct an experiment running the humor-detection task for four different embedding techniques (Word2Vec, GloVe, LSA, Dependency Weights) across three different setups: N-grams $+$ unweighted similarity scores ($S$), N-grams $+$ distance-weighted similarity scores ($WS$), N-grams $+$ unweighted similarity scores $+$ distance-weighted similarity scores ($S+WS$). The results presented below are exciting:

\begin{figure}[h]
\caption{Similarity/Discordance results across its twelve implemented experiments}
\centering
\includegraphics[width=0.5\textwidth]{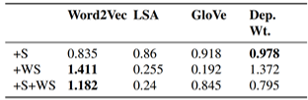}
\end{figure}

\begin{figure}[h]
\caption{Similarity/Discordance average increment in F1 performance across four common embedding techniques}
\centering
\includegraphics[width=0.5\textwidth]{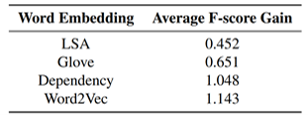}
\end{figure}

The approaches presented by \cite{Joshi} and \cite{Ghosh} are relatively intuitive, surprisingly effective, and allow for inference on the expected improvement in performance for humor-focused sentiment-analysis using additional contextualizing procedures. Yet, as previously mentioned, more powerful models can be built and yield stronger results by expanding this effort into the realm of neural architectures. The paper “A deep-learning framework to detect sarcasm targets” (\cite{Patro}) explores this concept, by including socio-linguistic features to augment a neural model's capacity to infer use of sarcasm from linguistic information beyond the word sequence. The proposed model uses the distribution of POS and NER tags (particularly LOC and ORG) alongside Empath – a tool to attribute a lexical category to a text’s topic – and the use of LIWC – used to analyze the positive emotions and negative emotions in the text. Additionally, as discussed in the work of “Sarcastic or Not: Word Embeddings to Predict the Literal or Sarcastic Meaning of Words” (\cite {Ghosh}), every word in the sentence is considered to be a possible target word. This is justified under the multiple candidate phrases hypothesis: in the sarcastic text “this laptop heats up so much that I strongly recommend chefs use it as a cooktop” the target candidates could be “chefs”, “cook-top” or “laptop” yet only “laptop” is ridiculed (\cite {Patro}).

The model architecture is elaborate, but remarkably interesting. A sentence with $n$ words is provided as input, and transformed into a tokenized sequence $\{w_1,…w_k…,w_n,<end>\}$. This sequence is fed into an input-layer of $n+1$ nodes where the $k^{th}$ node – corresponding to $w_k$ – is considered to be the target word. Each word is individually embedded using ELMo (\cite{elmo}) and passed in a one-to-one relation to the next layer. This next layer is experimented to be a LSTM, Bi-LSTM, or TD-LSTM layer, and is responsible for creating the context-embedding for the target word. The subsequence $\{ w_1,…,w_{k-1}\}$ is transformed into the context vector $c_{left}$, and the subsequence $\{w_{k+1},…,w_n,<end>\}$ yields $c_{right}$. Let the ELMo embedding of $w_k$ be $v_k$. The second layer then provides $\{c_{left},v_k,c_{right}\}$ onto the third layer, referred to as the concatenation layer. The concatenation layer will combine these three inputs, and depending on the experiment, it might perform the additional step of augmenting the concatenation with the socio-linguistic features previously discussed. Lastly the output of the concatenation layer is passed onto the dense layer, which applies a sigmoid activation function to perform the classification task for “sarcasm” or “not sarcasm”. The architecture can be visualized below with a diagram from the original paper.  

\begin{figure}[h]
\caption{The Deep Learning Framework (DLF) architecture as presented in the original paper}
\centering
\includegraphics[scale=0.55]{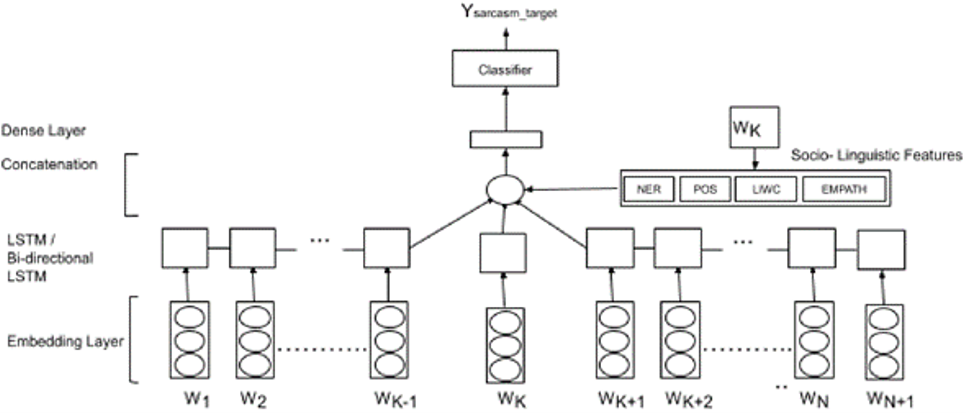}
\end{figure}

The performance of the proposed network is assessed across sixteen experiments, combining four designs for the second layer, two different datasets (tweets and snippets from public-domain books), and micro-level and macro-level metrics. $T$ and $S$ refer to the datasets, $\mu$ and $M$ refer to micro and macro-level analysis, and $slf$ refers to the socio-linguistic features. The results are presented below:

\begin{figure}[h]
\caption{F1 metric results achieved across sixteen experiments by the presented deep architecture}
\centering
\includegraphics[scale=0.8]{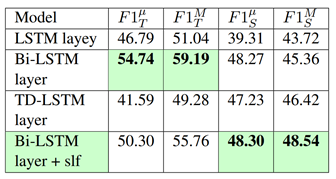}
\end{figure}

The results obtained by this paper are robust, and similarly to the two previous proposals analyzed, presents important progress on the first issue discussed in the introduction, regarding the need to design more context-aware models, capable of extracting clues to nonliteral meaning usage. However, none of these works have explored the second issue discussed, regarding the need to incorporate additional nonverbal information into contextualized embeddings. We introduce the paper “Words Can Shift: Dynamically Adjusting Word Representations Using Nonverbal Behaviors” by Wang et al. in which the authors argue for the importance of visual and acoustic information in the nonliteral meaning detection task. They state their goal as to better model multimodal human language incorporating nonverbal behaviors and learning multimodal-shifted word representations conditioned on nonverbal behaviors (\cite{wang}). They proposed the RAVEN model (Recurrent Attended Variation Embedding Network), which attempts to dynamically shift the embedding space of a word representation based on a one-to-one match between words in the sentence and visual and acoustic behavior associated with them. Particularly, the model seeks to identify clues as subtle as facial expressions and facial landmarks, and acoustic features such as pronunciation and pauses to learn shifted vectors that can disambiguate or emphasize the meaning of a word (\cite {wang}). Ultimately, the desire is to determine if the sequence of words and nonverbal information displays exact intent or nonliteral intent. 

The RAVEN model exhibits the most complex architecture between the analyzed models. Its central component is an attention mechanism which weights the verbal, acoustic, and visual components of the communication to produce a unified vector encapsulating multimodal information. This is relevant as for some training examples the visual input will be the most crucial for the detection, such as saying “I love this book” with a smile and relaxed facial muscles, and for others the acoustic information might be the most important, such as pronouncing the word “love” in “I love when my tea gets cold” very harshly. To handle these dynamic dependencies, the authors propose a gated attention mechanism that controls the importance of each visual and acoustic embedding for a word. The RAVEN architecture features three components:

\begin{enumerate}
    \item Nonverbal Sub-Networks model the nonverbal behavior using two independent LSTMs to encode a sequence of visual and acoustic patterns
    \item Gated Modality-Mixing Network takes as input the original word embedding and the visual and acoustic embeddings from the previous layer, and uses an attention mechanism to yield the shift vector in two possible directions: literal sense or nonliteral sense. It performs this task by learning word-specific nonlinear combinations of the three input embeddings
    \item Multimodal Shifting Layer computes the shift vector by modifying the original embedded vector according to the output of the attention layer. It does so by multiplying the respective visual and acoustic embeddings concatenated with the verbal embedding with corresponding weight matrices obtained from the attention layer
\end{enumerate}

\begin{figure}[h]
\caption{The RAVEN model architecture as presented in the original paper}
\centering
\includegraphics[scale=0.8]{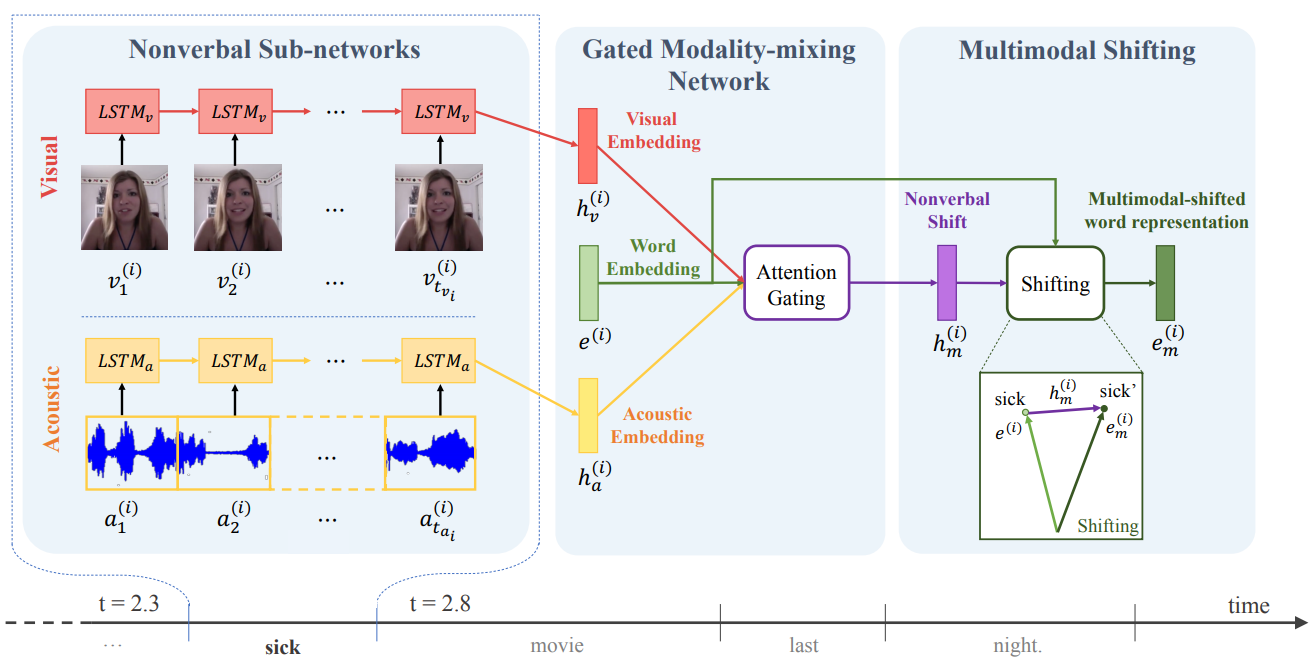}
\end{figure}

In short, the resulting shift vector $h_m$ is given by:
$$a_{vis}=\sigma(W_{vis} [h_{vis}\oplus h_{verb}]+b_{vis} )$$
$$a_{ac}=\sigma(W_{ac} [h_{ac}\oplus h_{verb}]+b_{ac} )$$
$$h_{shift}=a_{vis} \cdot (W_{vis}h_{vis}) + a_{ac} \cdot (W_{ac}h_{ac})+b_{shift}$$

Where $\sigma(\cdot)$ is the sigmoid activation function, $W_{vis},W_{ac}$ are the weight matrices from the attention step, $a_{vis},a_{ac}$ are the outputs of the attention gates, $h$ are the embedding vectors, and $b$ are the bias vectors.

Once $h_{shift}$ is obtained, the finalized combined verbal $+$ nonverbal embedding is given by:
$$h_{comb} = h_{verb} + \alpha \cdot h_{shift}$$

Where $\alpha=min\{\frac{||h_{verb}||}{||h_{shift}||}\beta, 1\}$ is a scaling factor prevents the shift vector from significantly outweighing the original verbal embedding. $\beta$ is a hyper-parameter. 

RAVEN was tested on the CMU-MOSI dataset (Multimodal Corpus of Sentiment Intensity $\&$ Subjectivity Analysis in Videos) constructed by Carnegie Mellon University. The dataset contains short videoclips of speakers demonstrating happiness, sadness, anger, or neutrality towards popular movies. RAVEN shows competitive performance against a set of competing models selected by the authors due to aligned motivations and goals. It is worth noting that RAVEN places in the top three for all metrics in the eight sentiment-analysis tasks explored among the nine analyzed architectures, which is a remarkable accomplishment as it demonstrates the model’s capacity to perform consistently and reliably. Results are reported below:

\begin{figure}[h]
\caption{Results achieved by RAVEN across different sentiment-tasks, compared to performance of current state-of-the-art implementations}
\centering
\includegraphics[scale=1]{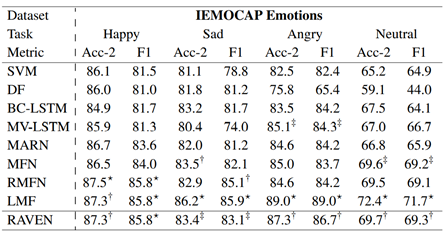}
\end{figure}

\section{Discussion on Results}

The four models presented and analysed by this survey were tested under different environments across widely different datasets. RAVEN (\cite {wang}) was tested on a multimodal dataset with videoclips, the LSSD classifier (\cite {Ghosh}) uses tweets labeled with ‘hashtags’ such as $\#$happy $\#$sad $\#$sarcasm, the similarity-based classifier (Similarity/Discordance) (\cite {Joshi}) uses user-posted book quotes extracted from an online forum, and the deep-learning framework (DLF) (\cite{Patro}) combines a broader collection of tweets alongside snippets from public-domain books. The datasets demonstrate vastly different sizes, topics, modes, and complexities, and unfortunately there is an additional complication due to papers not presenting their results in a standardized form. For these reasons, a direct comparison of F1 or accuracy performance is biased and unfair, and we will only report and analyze the metric improvements achieved by these models against their baselines.

First, it is important to note that all models exhibit performance improvements against their chosen baselines. RAVEN and LSSD can be most directly compared as they both utilize a variation of a SVM classifier as their baseline, most likely due to its reliability and computational simplicity in development. The mean improvement on F1 scores for RAVEN across all experiments is 4.32\% with a peak of 4.4\%, while LSSD has a mean improvement across experiments of 0.82\% with a peak of 1.4\%. This is not unexpected. RAVEN is a much more elaborate architecture leveraging on multimodal information extracted from nonverbal signals, while LSSD relies only on an additional statistical feature calculated at training-time. As none of the papers elaborate on the time and space complexity of training their implementations, we will refrain from discussing the efficiency trade-offs between complexity and performance, but any individual considering implementing one of these models should reflect upon the following question: is it worth it to allocate the additional time and resources to collect the massive amounts of highly-elaborate data required and train RAVEN, or will LSSD suffice for my needs?

The Similarity/Discordance classifier also uses a modified SVM classifier for its evaluation, however the baseline performance is not given by a SVM classifier, but rather through point-wise mutual information (PPMI) approach on the contextualized vectors. It is hard therefore to connect Similarity/Discordance to RAVEN and LSSD, so metric improvements reports must be taken cautiously as not to introduce biases in our judgement. The Similarity/Discordance classifier achieves F1 scores as much as 10.7\% higher than its baseline performance, however for some experiments the inclusion of the techniques described in the paper decreases the F1 score from the baseline. Consequently, the Similarity/Discordance classifier appears to be less reliable than its non-neural counterpart LSSD, as its variation in performance fluctuates greatly. The authors present a citation from the external paper “Improving distributional similarity with lessons learned from word embeddings” (Levy et al. [2015]) arguing that hyper-parametrization has a large impact on word embedding model success, so additional experimentation and proper hyper-parametrization might stabilize the model’s performance in the future. Currently, it should be considered less reliable than other models such as LSSD, and should be implemented carefully as it exhibits potential to over-fit.

Lastly, we try to connect these three models with the deep learning framework (DLF) for sarcasm detection. Intuitively, one could argue the DLF classifier can be directly compared with RAVEN due to their shared neural approaches, however DLF uses a neural baseline while RAVEN uses a SVM baseline, therefore improvements against baseline is a biased comparison. The DLF paper also only reports the baseline implementation's exact-match accuracy, not including the baseline's F1 scores. Only the experimentation of DLF with different hidden layers have reported F1 scores. In terms of exact-match accuracy, DLF performs best with a Bi-LSTM layer with the socio-linguistic approach included in the architecture, which is unsurprising since it produces forward and backwards context-vectors. For this setup it can outperform the baseline neural implementation by extremely outstanding 40.37\%. Unfortunately, it is currently unclear how this performance relates to RAVEN performance or any of the non-neural models, but still such a remarkable result should not be taken lightly.

In conclusion, LSSD can be considered a fast and reliable non-neural classifier for the nonliteral meaning detection task, but one should not expect performance on par with neural models. RAVEN requires very complex data to be trained and most likely requires the most computational resources out of all analyzed models to be developed as three independent networks must be trained simultaneously. However, if one is willing to bare the increased complexity of RAVEN opposed to LSSD, it will most likely lead to better performance. If employing RAVEN is not feasible due to its complex data requirements yet the performance of LSSD is not sufficient for a task, then DLF is a good alternative, placing itself as a middle point between them. Currently, there is not enough data to determine if DLF can outperform RAVEN, but considering that DLF is less complex in its architecture and requires less resources to be developed, one could also consider attempting DLF before RAVEN. If it is eventually empirically demonstrated by additional research that DLF exhibits competitive performance against RAVEN, then it could be considered a better alternative. Lastly, we note that Similarity/Discordance was too inconsistent on its results, and additional experiments need to be conducted before it can be more directly compared to other models.

\section{Conclusions}

The importance of advancing sentiment-analysis cannot be understated, so the efforts presented in this survey are extremely important to the field of Natural Language Processing, Robotics, and Artificial Intelligence, apart from pushing the boundaries of current research methods in Machine Learning and Deep Learning. Crucial conclusions have been drawn by recent experimentation in this field. Most importantly, it has become widely accepted that context and nonverbal communication are fundamental aspects of the way humans understand and express humor, so we must seek computational models capable of leveraging on these elements. To fully understand the nonliteral processes in the context of NLP, it is critical to consider the nature of the person who makes a statement, the nature of the person who receives it, and the context in which this social interaction occurs (\cite {Katz}).

Improving these language models opens remarkable opportunities to the possibilities presented in the introduction, most noticeably social robots capable of not only expressing humor, but also understanding it and adapting accordingly to increase their likeability, effectiveness, and acceptance (\cite {Ritschel}). There are exciting opportunities ahead combining these models with other artificial intelligence techniques to develop social robots with personalities and adaptive behavior towards users and patients. Work in this area is already being conducted as demonstrated by the paper “Shaping a Social Robot’s Humor with Natural Language Generation and Socially-Aware Reinforcement Learning” (\cite {Ritschel}). The goal is to work towards social robots using reinforcement learning to adapt their personality to individual users’ preferences by reacting to user behavior and adapting conduct without the need of scripted reactions (\cite {Ritschel}). It is undeniable that models such as RAVEN would play an monumental role in this context, as Professor Ritschel and his colleagues describe an ideal approach using human social signals, primarily vocal laughter and smile to estimate the spectator’s reaction to the robot, which would be used as the reward parameter to the reinforcement learning approach.

Many challenges still lay ahead. First it is relevant to warn individuals interested in furthering this research that collecting data for such studies is very challenging and expensive. All models presented in this paper employ supervised learning at some step of development, and annotated real-world sarcasm data is not readily available. Researchers have resorted to crowdsourcing platforms such as Amazon Mechanical Turk, paying volunteer annotators to analyze and label text snippets such as tweets and book quotes. Not only this process is slow and expensive, but it can also be unreliable and inconsistent. Common criticism of research based on use of tweets and hashtags as gold labels is that the training utterances could be noisy (\cite {Ghosh}), meaning tweets may or may not be sarcastic without presenting clear evidence, and annotators analyzing the tweets without context and nonverbal clues could make mistakes. This is of course a lesser problem for a model like RAVEN, but collecting data for RAVEN is an even harder task because it cannot be simply scrapped from websites as it requires recording videoclips of people speaking.

For this reason, future work in this area will most likely see attempts to reduce the dependability on enormous amounts of annotated data by implementing unsupervised learning steps. A potential approach can be drawn inspired by data debiasing work applying embedding projections onto “neutral” vector-spaces. It is worth exploring the same concept for “humorous” vector-spaces and “literal meaning” vector-spaces projections, attempting to relate word-vector embeddings to these vector-spaces. I believe that regardless of future directions, this survey has presented a representative overview of current techniques being applied to the humor-detection task and what successes have been achieved, giving a good starting point to any researcher interested in expanding the state-of-the-art in humor-focused sentiment-analysis. 

\newpage

\bibliography{references}

\end{document}